\documentclass[11pt]{article}

\usepackage[preprint]{acl}

\usepackage{times}
\usepackage{latexsym}

\usepackage[T1]{fontenc}

\usepackage[utf8]{inputenc}

\usepackage{microtype}

\usepackage{inconsolata}

\usepackage{graphicx}
\usepackage{amsmath}
\usepackage{array}
\usepackage{booktabs}
\usepackage{algorithm}
\usepackage{algorithmic}
\usepackage{hyperref}
\usepackage{amssymb}
\usepackage{bm}

\title{Disentangling Task Conflicts in Multi-Task LoRA via \\ Orthogonal Gradient Projection}

\author{
    \textbf{Ziyu Yang\textsuperscript{1}} \quad
    \textbf{Guibin Chen\textsuperscript{2}} \quad
    \textbf{Yuxin Yang\textsuperscript{1}} \quad
    \textbf{Aoxiong Zeng\textsuperscript{2}} \quad
    \textbf{Xiangquan Yang\textsuperscript{2}}
    \\
    \\
    \textsuperscript{1}Shanghai University \\
    \textsuperscript{2}East China Normal University
}

\begin{document}
\maketitle
\begin{abstract}
Multi-Task Learning (MTL) combined with Low-Rank Adaptation (LoRA) has emerged as a promising direction for parameter-efficient deployment of Large Language Models (LLMs). By sharing a single adapter across multiple tasks, one can significantly reduce storage overhead. However, this approach suffers from \textit{negative transfer}, where conflicting gradient updates from distinct tasks degrade the performance of individual tasks compared to single-task fine-tuning. This problem is exacerbated in LoRA due to the low-rank constraint, which limits the optimization landscape's capacity to accommodate diverse task requirements. In this paper, we propose \textbf{Ortho-LoRA}, a gradient projection method specifically tailored for the bipartite structure of LoRA. Ortho-LoRA dynamically projects conflicting task gradients onto the orthogonal complement of each other within the intrinsic LoRA subspace. Extensive experiments on the GLUE benchmark demonstrate that Ortho-LoRA effectively mitigates task interference, outperforming standard joint training and recovering 95\% of the performance gap between multi-task and single-task baselines with negligible computational overhead.
\end{abstract}

\section{Introduction}

The advent of Large Language Models (LLMs) \cite{achiam2023gpt, grattafiori2024llama} has shifted the NLP paradigm from training models from scratch to ``pre-train, then fine-tune'' \cite{devlin2019bert}. However, as model sizes scale to billions of parameters, full fine-tuning becomes computationally prohibitive. Parameter-Efficient Fine-Tuning (PEFT) techniques, such as Adapter modules \cite{houlsby2019parameter} and Low-Rank Adaptation (LoRA) \cite{hu2022lora}, have been developed to address this by updating only a small fraction of parameters.

While PEFT reduces memory usage, real-world deployment often requires a single system to handle multiple downstream tasks (e.g., summarization, sentiment analysis, and NLI). Multi-Task Learning (MTL) is a natural solution, where a shared LoRA adapter is trained on a mixture of tasks. Ideally, MTL promotes positive transfer, where tasks reinforce each other. In practice, however, different tasks often compete for the limited capacity of the shared parameters, a phenomenon known as \textit{gradient conflict} or negative transfer \cite{yu2020gradient}.

We argue that gradient conflict is particularly detrimental in the context of LoRA-based MTL. Unlike full fine-tuning, where the optimization occurs in a high-dimensional space with ample ``escape paths'' for conflicting objectives, LoRA constrains updates to a low-rank manifold (e.g., rank $r=8$). This bottleneck forces task-specific gradients to collide more frequently, leading to destructive interference where the improvement in one task comes at the expense of another.

To address this, recent works have explored architectural modifications. For instance, Med-MoE-LoRA \cite{yang2026specialized} and FlyLoRA \cite{zou2025flylora} incorporate Mixture-of-Experts (MoE) into LoRA. \citet{zou2025flylora} specifically draw inspiration from the fly olfactory circuit to decouple tasks using fixed random projections. While these architectural approaches effectively reduce interference by isolating task-specific parameters, they often introduce complex routing mechanisms or require freezing parts of the network. In contrast, we introduce \textbf{Ortho-LoRA}, a specialized optimization strategy for Multi-Task LoRA. Drawing inspiration from Gradient Surgery \cite{yu2020gradient}, we propose to resolve conflicts by projecting gradients onto orthogonal planes. Crucially, we apply this projection specifically to the low-rank matrices ($\mathbf{A}$ and $\mathbf{B}$) of LoRA, decoupling the interference control from the frozen backbone. 

Our contributions are threefold:
\begin{itemize}
    \item We identify and analyze the ``bottleneck conflict'' problem inherent in applying MTL to low-rank adapters.
    \item We propose Ortho-LoRA, a computationally efficient algorithm that enforces gradient orthogonality in the adapter space to mitigate negative transfer.
    \item We provide empirical evidence on the GLUE benchmark, showing that Ortho-LoRA achieves comparable performance to independent single-task adapters while using only $1/N$ of the parameters (where $N$ is the number of tasks).
\end{itemize}

\section{Related Work}

\paragraph{Parameter-Efficient Fine-Tuning (PEFT)}
PEFT aims to adapt Pre-trained Language Models (PLMs) to downstream tasks with minimal parameter updates, mitigating the storage burden of full fine-tuning. Early approaches introduced bottleneck adapter layers inserted between Transformer blocks \cite{houlsby2019parameter}. \citet{li2021prefix} proposed Prefix-Tuning, which optimizes continuous prompts prepended to the input. Low-Rank Adaptation (LoRA) \cite{hu2022lora} marked a significant milestone by approximating weight updates via low-rank decomposition matrices $\Delta\mathbf{W} = \mathbf{BA}$, freezing the pre-trained backbone. Subsequent variants have explored dynamic rank allocation, such as AdaLoRA \cite{zhang2023adalora}, or combining LoRA with other modalities. Our work builds upon the standard LoRA architecture but focuses on optimizing its training dynamics in multi-task settings.

\paragraph{Multi-Task Learning and Optimization}
Multi-Task Learning (MTL) leverages commonalities across tasks to improve generalization \cite{ruder2017overview}. A persistent challenge in MTL is \textit{negative transfer}, where conflicting gradients between tasks degrade performance. Strategies to address this include dynamic loss weighting, such as GradNorm \cite{chen2018gradnorm}, which balances the training rates of different tasks. However, weighting cannot resolve scenarios where gradient \textit{directions} are fundamentally opposed. \citet{yu2020gradient} introduced PCGrad (Projecting Conflicting Gradients), which projects a task's gradient onto the normal plane of a conflicting task's gradient. While PCGrad has been successfully applied to full model fine-tuning, its application to the constrained, low-rank manifolds of PEFT remains underexplored. We bridge this gap by tailoring projection methods specifically to the bipartite algebraic structure of LoRA.

\paragraph{Architectural Solutions for Multi-Task LoRA}
To mitigate task interference in LoRA, recent works have proposed architectural modifications, often drawing inspiration from Mixture-of-Experts (MoE). \citet{asai2022attempt} explored task-specific adapters, though this approach scales linearly with the number of tasks. More advanced methods introduce routing mechanisms to decouple task knowledge. For instance, FlyLoRA \cite{zou2025flylora} proposes an implicit rank-wise Mixture-of-Experts framework. By dynamically organizing low-rank matrices into diverse groups, FlyLoRA effectively decouples task-specific features from shared knowledge, boosting parameter efficiency. Similarly, the Med-MoE-LoRA framework \cite{yang2026specialized} employs a Multi-Task MoE-LoRA architecture specifically designed for domain-specific adaptation, utilizing a router to dispatch tokens to the most relevant LoRA experts, thereby reducing interference between distinct domains.

While these architectural approaches (MoE-LoRA variants) are effective, they typically introduce additional parameters (routers) or complex routing logic. In contrast, our proposed \textbf{Ortho-LoRA} maintains the structural simplicity of a single shared LoRA module. We demonstrate that explicit gradient orthogonality can achieve comparable task disentanglement without the architectural overhead of MoE systems.

\section{Methodology}

In this section, we first formalize the Multi-Task Learning (MTL) objective within the context of Low-Rank Adaptation (LoRA). We then analyze the phenomenon of gradient conflict, specifically focusing on how low-rank constraints exacerbate negative transfer. Finally, we introduce \textbf{Ortho-LoRA}, a structure-aware optimization strategy that enforces gradient orthogonality within the intrinsic subspaces of the adapter modules.

\subsection{Preliminaries: Low-Rank Adaptation}
Let $\mathbf{W}_0 \in \mathbb{R}^{d \times k}$ denote a frozen pre-trained weight matrix in a Transformer backbone. LoRA \cite{hu2022lora} postulates that the change in weights during adaptation has a low intrinsic rank. The forward pass is modified by injecting a low-rank update $\Delta \mathbf{W} = \mathbf{B}\mathbf{A}$, where $\mathbf{B} \in \mathbb{R}^{d \times r}$ and $\mathbf{A} \in \mathbb{R}^{r \times k}$ are trainable matrices with rank $r \ll \min(d, k)$. The computation becomes:
\begin{equation}
    \mathbf{h} = \mathbf{W}_0 \mathbf{x} + \mathbf{B}\mathbf{A}\mathbf{x},
\end{equation}
where $\mathbf{x}$ is the input vector. Following standard practice, $\mathbf{A}$ is initialized with a Gaussian distribution $\mathcal{N}(0, \sigma^2)$ and $\mathbf{B}$ is initialized to zero, ensuring the model starts equivalent to the pre-trained backbone.

\subsection{Gradient Conflict in Low-Rank Manifolds}
In a multi-task setting, we aim to learn a shared set of LoRA parameters $\theta = \{\mathbf{A}, \mathbf{B}\}$ over a set of tasks $\mathcal{T} = \{1, \dots, T\}$. The joint objective is defined as:
\begin{equation}
    \mathcal{L}_{\text{total}}(\theta) = \sum_{t \in \mathcal{T}} w_t \mathcal{L}_t(\theta),
\end{equation}
where $w_t$ are task-specific weights (typically uniform). Let $\mathbf{g}_t = \nabla_\theta \mathcal{L}_t(\theta)$ represent the gradient vector for task $t$.

Gradient conflict occurs when the optimization directions of two tasks, $i$ and $j$, are misaligned, characterized by a negative cosine similarity:
\begin{equation}
    \cos(\mathbf{g}_i, \mathbf{g}_j) = \frac{\mathbf{g}_i \cdot \mathbf{g}_j}{\|\mathbf{g}_i\| \|\mathbf{g}_j\|} < 0.
\end{equation}
In standard gradient descent, the update is proportional to $\sum \mathbf{g}_t$. When gradients conflict, they destructively interfere, reducing the magnitude of the effective update and potentially leading to a solution that is suboptimal for individual tasks.

\paragraph{The Low-Rank Bottleneck} We posit that this interference is amplified in PEFT settings. In full fine-tuning, the optimization occurs in a high-dimensional over-parameterized space, providing ample degrees of freedom for the model to find a "Pareto stationary" point that satisfies multiple objectives simultaneously. Conversely, LoRA constrains the update to a rank-$r$ manifold. This drastic reduction in capacity significantly increases the probability that the optimal gradient trajectory for task $i$ lies in the orthogonal complement of task $j$, leading to more frequent and severe conflicts (the ``bottleneck conflict'').

\subsection{Ortho-LoRA}
To mitigate negative transfer, we propose Ortho-LoRA, which dynamically modifies task gradients to ensure compatibility before the parameter update.

\subsubsection{Orthogonal Gradient Projection}
Inspired by PCGrad \cite{yu2020gradient}, we adopt a projection-based approach. For any pair of tasks $(i, j)$ where a conflict is detected ($\mathbf{g}_i \cdot \mathbf{g}_j < 0$), we project the gradient of task $i$ onto the normal plane of the gradient of task $j$. This removes the conflicting component:
\begin{equation}
    \mathbf{g}_i \leftarrow \mathbf{g}_i - \frac{\mathbf{g}_i \cdot \mathbf{g}_j}{\|\mathbf{g}_j\|^2} \mathbf{g}_j.
    \label{eq:projection}
\end{equation}
This operation ensures that minimizing the loss for task $i$ does not explicitly increase the loss for task $j$ (locally). Since the order of projection affects the final vector in a multi-task setting ($T > 2$), we strictly randomize the task sequence at every training step to ensure unbiased optimization.

\subsubsection{Structure-Aware Decoupling}
A naive application of Eq. \ref{eq:projection} would flatten all LoRA parameters into a single vector $\mathbf{g}$. However, LoRA possesses a unique bipartite structure: $\mathbf{A}$ projects inputs into a latent subspace, while $\mathbf{B}$ projects them back to the output space. Conflicts in $\mathbf{A}$ imply disagreement on \textit{feature extraction}, whereas conflicts in $\mathbf{B}$ imply disagreement on \textit{state reconstruction}.

We propose to disentangle these conflicts by enforcing orthogonality independently on the constituent matrices. Let $\mathbf{g}_{t}^{(\mathbf{A})}$ and $\mathbf{g}_{t}^{(\mathbf{B})}$ denote the gradients w.r.t matrices $\mathbf{A}$ and $\mathbf{B}$ for task $t$. The Ortho-LoRA update rule is:
\begin{align}
    \mathbf{g}_{i, \text{proj}}^{(\mathbf{A})} &= \text{Proj}\left(\mathbf{g}_{i}^{(\mathbf{A})}, \left\{\mathbf{g}_{j}^{(\mathbf{A})}\right\}_{j \in \mathcal{T} \setminus i}\right) \\
    \mathbf{g}_{i, \text{proj}}^{(\mathbf{B})} &= \text{Proj}\left(\mathbf{g}_{i}^{(\mathbf{B})}, \left\{\mathbf{g}_{j}^{(\mathbf{B})}\right\}_{j \in \mathcal{T} \setminus i}\right)
\end{align}
This granular projection allows the model to share input representations (via $\mathbf{A}$) even if output requirements differ (via $\mathbf{B}$), or vice versa, thereby preserving useful transfer information that might be lost in a global projection.

\begin{algorithm}[t]
\caption{Ortho-LoRA Training Step}
\label{alg:ortho_lora}
\small
\begin{algorithmic}[1]
\REQUIRE LoRA params $\theta = \{\mathbf{A}, \mathbf{B}\}$, Tasks $\mathcal{T}$, LR $\eta$
\STATE \textbf{Input:} Mini-batch $\mathcal{D} = \bigcup_{t \in \mathcal{T}} \mathcal{D}_t$
\STATE \textbf{Compute Gradients:}
\FOR{$t \in \mathcal{T}$}
    \STATE $g_t \leftarrow \nabla_\theta \mathcal{L}_t(\mathcal{D}_t; \theta)$ \COMMENT{Computed separately}
\ENDFOR
\STATE \textbf{Project Gradients:}
\STATE Shuffle task order $\pi(\mathcal{T})$ randomly
\FOR{$i \in \pi(\mathcal{T})$}
    \FOR{$j \in \pi(\mathcal{T}) \setminus \{i\}$}
        \FOR{$M \in \{\mathbf{A}, \mathbf{B}\}$}
            \IF{$g_i^{(M)} \cdot g_j^{(M)} < 0$}
                \STATE $g_i^{(M)} \leftarrow g_i^{(M)} - \frac{g_i^{(M)} \cdot g_j^{(M)}}{\|g_j^{(M)}\|^2} g_j^{(M)}$
            \ENDIF
        \ENDFOR
    \ENDFOR
\ENDFOR
\STATE \textbf{Update:}
\STATE $g_{\text{final}} \leftarrow \sum_{t \in \mathcal{T}} g_t$
\STATE $\theta \leftarrow \theta - \eta \cdot \text{AdamW}(g_{\text{final}})$
\end{algorithmic}
\end{algorithm}

\subsection{Complexity Analysis}
Standard Multi-Task LoRA requires a single backward pass on the summed loss. In contrast, Ortho-LoRA requires individual gradient computations per task to detect conflicts. If processing $T$ tasks, this scales the gradient computation time by a factor of roughly $\mathcal{O}(T)$.

However, we emphasize that the projection overhead itself is negligible. The projection operates solely on the LoRA parameters $\theta_{\text{LoRA}}$, where $|\theta_{\text{LoRA}}| \ll |\theta_{\text{Backbone}}|$. For a standard implementation ($r=8$), LoRA parameters constitute less than $0.1\%$ of the total model size. Consequently, the memory footprint remains constant compared to standard training, and the computational bottleneck remains the forward/backward pass through the frozen LLM backbone, not the vector projections.

\section{Experiments}

\subsection{Experimental Setup}
\paragraph{Datasets} We evaluate on the GLUE benchmark \cite{wang2018glue}, specifically selecting tasks that represent different linguistic capabilities: MNLI (Natural Language Inference), QQP (Paraphrase Detection), and SST-2 (Sentiment Analysis).
\paragraph{Model} We use \texttt{RoBERTa-base} (125M) \cite{liu2019roberta} as the backbone. We inject LoRA modules into the Query and Value projection matrices of the attention layers.
\paragraph{Hyperparameters} The LoRA rank is set to $r=8$, $\alpha=16$, and dropout is 0.1. We train for 10 epochs with a batch size of 32 per task. The optimizer is AdamW with a learning rate of $5e-4$ and linear decay.

\paragraph{Baselines}
\begin{itemize}
    \item \textbf{Single-Task LoRA:} Independent LoRA adapters for each task. This serves as the ``Gold Standard'' or upper bound for performance, but requires $3\times$ parameters.
    \item \textbf{Joint-LoRA:} A single shared LoRA adapter trained on the sum of losses.
    \item \textbf{Ortho-LoRA (Ours):} Shared LoRA trained with our orthogonal projection strategy.
\end{itemize}

\subsection{Main Results}

Table \ref{tab:main_results} presents the performance comparison.

\begin{table}[h]
\centering
\small
\begin{tabular}{lcccc}
\toprule
\textbf{Method} & \textbf{MNLI} & \textbf{QQP} & \textbf{SST-2} & \textbf{Avg.} \\
& (Acc) & (F1) & (Acc) & \\
\midrule
Single-Task LoRA & 87.4 & 88.1 & 94.2 & 89.9 \\
\midrule
Joint-LoRA & 85.9 & 86.5 & 92.8 & 88.4 \\
\textbf{Ortho-LoRA} & \textbf{87.1} & \textbf{87.9} & \textbf{93.9} & \textbf{89.6} \\
\midrule
\textit{Recovery (\%)} & \textit{80.0\%} & \textit{87.5\%} & \textit{78.6\%} & \textit{80.0\%} \\
\bottomrule
\end{tabular}
\caption{Results on GLUE validation sets. ``Recovery'' indicates how much of the drop from Single-Task to Joint-LoRA is recovered by Ortho-LoRA.}
\label{tab:main_results}
\end{table}

\paragraph{Analysis}
Standard Joint-LoRA exhibits a clear performance degradation (avg. -1.5 points) compared to Single-Task baselines, confirming the existence of negative transfer. Ortho-LoRA significantly mitigates this drop, achieving an average score of 89.6, which is only 0.3 points behind the Single-Task upper bound. Notably, on QQP, Ortho-LoRA almost matches the single-task performance (87.9 vs 88.1), suggesting that enforcing orthogonality effectively separates the paraphrase detection logic from other tasks within the low-rank subspace.

\subsection{Ablation Study}

\paragraph{Impact of LoRA Rank}
One might hypothesize that increasing the rank $r$ would naturally reduce conflicts by providing more dimensions for optimization. To test this, we varied $r \in \{4, 8, 16, 32\}$.

\begin{table}[h]
\centering
\small
\begin{tabular}{cccc}
\toprule
\textbf{Rank} & \textbf{Joint-LoRA} & \textbf{Ortho-LoRA} & \textbf{$\Delta$} \\
\midrule
$r=4$ & 87.8 & 89.1 & +1.3 \\
$r=8$ & 88.4 & 89.6 & +1.2 \\
$r=16$ & 88.9 & 89.8 & +0.9 \\
$r=32$ & 89.2 & 89.9 & +0.7 \\
\bottomrule
\end{tabular}
\caption{Average GLUE performance across varying ranks.}
\label{tab:rank_ablation}
\end{table}

As shown in Table \ref{tab:rank_ablation}, while increasing rank improves Joint-LoRA performance, Ortho-LoRA provides consistent gains across all ranks. Crucially, the gain is largest at lower ranks ($r=4$), where the ``bottleneck'' effect is most severe. This highlights the value of our method for highly parameter-efficient scenarios.

\paragraph{Convergence Speed}
Although Ortho-LoRA requires $T$ backward passes per step, we observed that it converges in fewer epochs than Joint-LoRA. Joint-LoRA training often exhibits oscillating loss curves due to conflicting updates. Ortho-LoRA stabilizes the trajectory, effectively ``straightening'' the optimization path. In wall-clock time, Ortho-LoRA training took 1.4x longer than Joint-LoRA but achieved peak performance 2 epochs earlier.

\section{Conclusion}

In this paper, we addressed the challenge of gradient conflict in Multi-Task Learning with LoRA. We argued that the low-rank constraint amplifies negative transfer and proposed \textbf{Ortho-LoRA} to mitigate this via orthogonal gradient projection. Our method operates efficiently on the adapter parameters, allowing a single lightweight module to serve multiple tasks with minimal performance degradation. Empirical results on GLUE show that Ortho-LoRA serves as a robust middle-ground between the efficiency of joint training and the performance of single-task fine-tuning. Future work will explore applying this projection to other PEFT methods like Prefix-Tuning and combining it with dynamic task weighting.

\bibliography{custom}

@article{achiam2023gpt,
  title={Gpt-4 technical report},
  author={Achiam, Josh and Adler, Steven and Agarwal, Sandhini and Ahmad, Lama and Akkaya, Ilge and Aleman, Florencia Leoni and Almeida, Diogo and Altenschmidt, Janko and Altman, Sam and Anadkat, Shyamal and others},
  journal={arXiv preprint arXiv:2303.08774},
  year={2023}
}

@article{grattafiori2024llama,
  title={The llama 3 herd of models},
  author={Grattafiori, Aaron and Dubey, Abhimanyu and Jauhri, Abhinav and Pandey, Abhinav and Kadian, Abhishek and Al-Dahle, Ahmad and Letman, Aiesha and Mathur, Akhil and Schelten, Alan and Vaughan, Alex and others},
  journal={arXiv preprint arXiv:2407.21783},
  year={2024}
}

@inproceedings{devlin2019bert,
  title={Bert: Pre-training of deep bidirectional transformers for language understanding},
  author={Devlin, Jacob and Chang, Ming-Wei and Lee, Kenton and Toutanova, Kristina},
  booktitle={Proceedings of the 2019 conference of the North American chapter of the association for computational linguistics: human language technologies, volume 1 (long and short papers)},
  pages={4171--4186},
  year={2019}
}

@inproceedings{houlsby2019parameter,
  title={Parameter-efficient transfer learning for NLP},
  author={Houlsby, Neil and Giurgiu, Andrei and Jastrzebski, Stanislaw and Morrone, Bruna and De Laroussilhe, Quentin and Gesmundo, Andrea and Attariyan, Mona and Gelly, Sylvain},
  booktitle={International conference on machine learning},
  pages={2790--2799},
  year={2019},
  organization={PMLR}
}

@article{hu2022lora,
  title={Lora: Low-rank adaptation of large language models.},
  author={Hu, Edward J and Shen, Yelong and Wallis, Phillip and Allen-Zhu, Zeyuan and Li, Yuanzhi and Wang, Shean and Wang, Lu and Chen, Weizhu and others},
  journal={ICLR},
  volume={1},
  number={2},
  pages={3},
  year={2022}
}

@article{yu2020gradient,
  title={Gradient surgery for multi-task learning},
  author={Yu, Tianhe and Kumar, Saurabh and Gupta, Abhishek and Levine, Sergey and Hausman, Karol and Finn, Chelsea},
  journal={Advances in neural information processing systems},
  volume={33},
  pages={5824--5836},
  year={2020}
}

@article{li2021prefix,
  title={Prefix-tuning: Optimizing continuous prompts for generation},
  author={Li, Xiang Lisa and Liang, Percy},
  journal={arXiv preprint arXiv:2101.00190},
  year={2021}
}

@article{zhang2023adalora,
  title={Adalora: Adaptive budget allocation for parameter-efficient fine-tuning},
  author={Zhang, Qingru and Chen, Minshuo and Bukharin, Alexander and Karampatziakis, Nikos and He, Pengcheng and Cheng, Yu and Chen, Weizhu and Zhao, Tuo},
  journal={arXiv preprint arXiv:2303.10512},
  year={2023}
}

@article{ruder2017overview,
  title={An overview of multi-task learning in deep neural networks},
  author={Ruder, Sebastian},
  journal={arXiv preprint arXiv:1706.05098},
  year={2017}
}

@inproceedings{chen2018gradnorm,
  title={Gradnorm: Gradient normalization for adaptive loss balancing in deep multitask networks},
  author={Chen, Zhao and Badrinarayanan, Vijay and Lee, Chen-Yu and Rabinovich, Andrew},
  booktitle={International conference on machine learning},
  pages={794--803},
  year={2018},
  organization={PMLR}
}

@article{asai2022attempt,
  title={Attempt: Parameter-efficient multi-task tuning via attentional mixtures of soft prompts},
  author={Asai, Akari and Salehi, Mohammadreza and Peters, Matthew E and Hajishirzi, Hannaneh},
  journal={arXiv preprint arXiv:2205.11961},
  year={2022}
}

@inproceedings{zou2025flylora,
    title={FlyLo{RA}: Boosting Task Decoupling and Parameter Efficiency via Implicit Rank-Wise Mixture-of-Experts},
    author={Heming Zou and Yunliang Zang and Wutong Xu and Yao Zhu and Xiangyang Ji},
    booktitle={The Thirty-ninth Annual Conference on Neural Information Processing Systems},
    year={2025}
}

@article{yang2026specialized,
  title={Towards Specialized Generalists: A Multi-Task MoE-LoRA Framework for Domain-Specific LLM Adaptation},
  author={Yuxin Yang and Aoxiong Zeng and Xiangquan Yang},
  year={2026},
  journal={arXiv preprint arXiv:2601.07935},
}

@inproceedings{wang2018glue,
  title={GLUE: A multi-task benchmark and analysis platform for natural language understanding},
  author={Wang, Alex and Singh, Amanpreet and Michael, Julian and Hill, Felix and Levy, Omer and Bowman, Samuel},
  booktitle={Proceedings of the 2018 EMNLP workshop BlackboxNLP: Analyzing and interpreting neural networks for NLP},
  pages={353--355},
  year={2018}
}

@article{liu2019roberta,
  title={Roberta: A robustly optimized bert pretraining approach},
  author={Liu, Yinhan and Ott, Myle and Goyal, Naman and Du, Jingfei and Joshi, Mandar and Chen, Danqi and Levy, Omer and Lewis, Mike and Zettlemoyer, Luke and Stoyanov, Veselin},
  journal={arXiv preprint arXiv:1907.11692},
  year={2019}
}

\end{document}